\documentclass{article}
 \PassOptionsToPackage{numbers, compress}{natbib}

 \usepackage[preprint]{nips_2018}

\usepackage[utf8]{inputenc} 
\usepackage[T1]{fontenc}    
\usepackage{hyperref}       
\usepackage{url}            
\usepackage{booktabs}       
\usepackage{amsfonts}       
\usepackage{nicefrac}       
\usepackage{microtype}      
\usepackage{times}
\usepackage{epsfig}
\usepackage{graphicx}
\usepackage{amsmath}
\usepackage{amssymb}
\usepackage{algorithm}
\usepackage{algorithmic}
\usepackage{mathrsfs}
\usepackage{array}
\usepackage{bm}
\usepackage{multirow}
\usepackage{subfigure}
\usepackage{wrapfig}

\title{Deep LDA Hashing}

\author{Di~Hu\\
        School of Computer Science\\
        Northwestern Polytechnical University\\
        \And
        Feiping Nie\\
        School of Computer Science\\
        Northwestern Polytechnical University\\
        \And
        Xuelong~Li\\
        Chinese Academy of Sciences
}

\begin{document}

\maketitle

\begin{abstract}
The conventional supervised hashing methods based on classification do not entirely meet the requirements of hashing technique, but \emph{Linear Discriminant Analysis} (LDA) does.
In this paper, we propose to perform a revised LDA objective over deep networks to learn efficient hashing codes in a truly end-to-end fashion.
However, the complicated eigenvalue decomposition within each mini-batch in every epoch has to be faced with when simply optimizing the deep network w.r.t. the LDA objective.
In this work, the revised LDA objective is transformed into a simple least square problem, which naturally overcomes the intractable problems and can be easily solved by the off-the-shelf optimizer.
Such deep extension can also overcome the weakness of LDA Hashing in the limited linear projection and feature learning.
Amounts of experiments are conducted on three benchmark datasets. The proposed Deep LDA Hashing shows nearly 70 points improvement over the conventional one on the CIFAR-10 dataset.
It also beats several state-of-the-art methods on various metrics.
\end{abstract}

\section{Introduction}
In the era of big data, the amount of data is growing rapidly along with the popularization of multiple kinds of digital recording devices, especially the visual, audio, and text data.
To satisfy the required huge storage space, and organization, learning capacity in dealing with such big data, hashing technique has been widely employed to learn effective binary representation in multiple tasks~\cite{kafai2014discrete,zhang2015bit,li2017deep}, especially the image retrieval task~\cite{li2017large}. This is because the binary representations take advantage of the coding property of existing digital recording manner, i.e., we can take one Byte to encode one specific image of several MBytes.

Hashing is mainly developed to map the image or document into short binary code sequence while preserving the similarity structure among the original data.
Most works focus on the data-independent methods in the earlier studies, as such methods can enjoy low computation complexity via random projection.
The most representative work is \emph{Locality Sensitive Hashing} (LSH)~\cite{andoni2006near}. Several extensions based on LSH are also proposed, e.g., kernel LSH~\cite{kulis2012kernelized} and $p$-norm LSH~\cite{datar2004locality}.
However, the obtained hashing codes suffer from the inefficiency of random projection, which therefore requires longer codes to achieve better performance~\cite{liu2014discrete}.

To generate more compact codes, more attention are turned on to the data-dependent methods. Different from the random projection, these methods propose to learn effective hashing functions via exploring the intrinsic data structure or depending on the labels. Hence there are two major categories among them, one is unsupervised hashing, and the other is supervised hashing.
Specifically, the former focuses on how to maintain the data structure and make the generated codes more informative. For example, Iterative Quantization (ITQ)~\cite{gong2013iterative} adopts PCA projection to maximize the variance of hash functions, while Isotropic Hashing~\cite{kong2012isotropic} targets to make the projected dimensions into equal variance. The reconstruction strategy is also considered to generate effective low-dimensional representation~\cite{kulis2009learning,carreira2015hashing}. In addition, the Laplacian eigenmap is employed to learn the intrinsic manifold structure and nonlinearly embed the data into proper codes, e.g., Spectral Hashing (SH)~\cite{weiss2009spectral}, spectral rotation~\cite{li2017large}.

\begin{wrapfigure}{R}{0.5\textwidth}
\begin{center}
  \subfigure[Classification]{
    \label{aa}
    \includegraphics[width=0.23\textwidth]{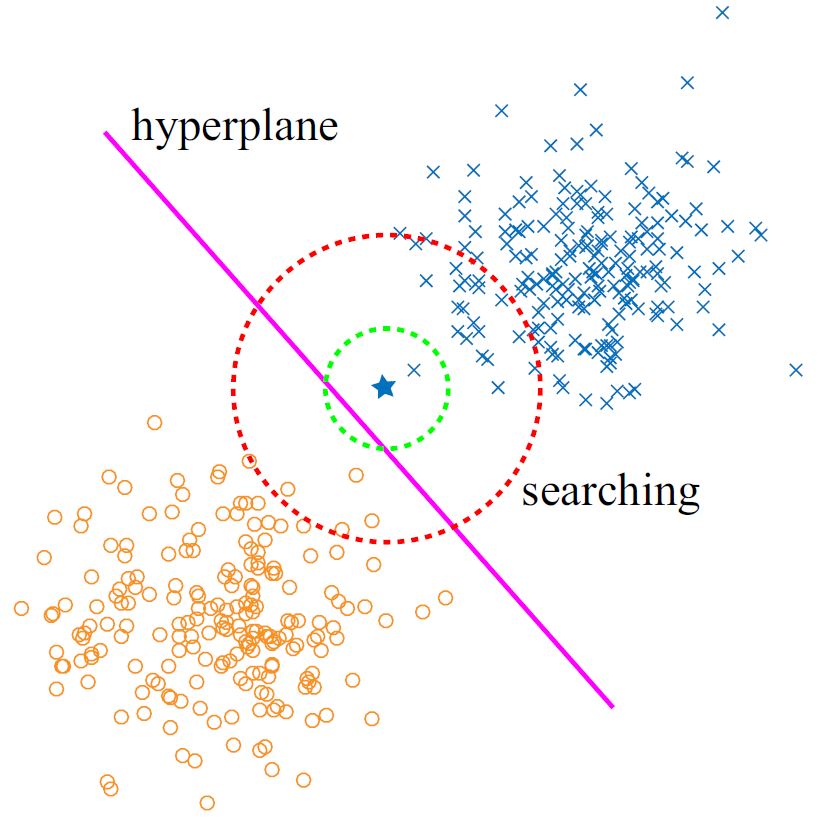}}
  \subfigure[LDA]{
    \label{bb}
    \includegraphics[width=0.23\textwidth]{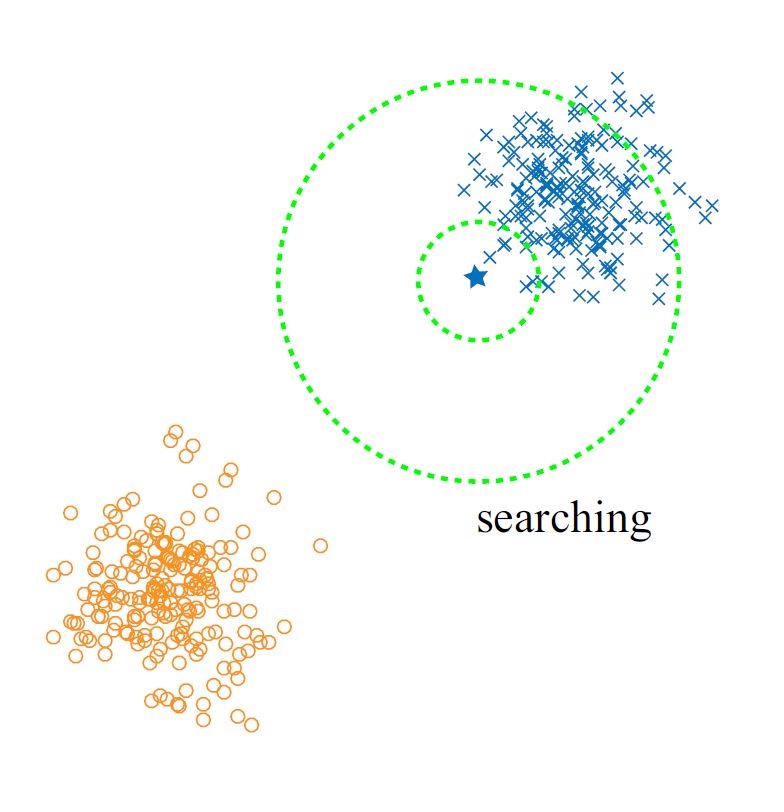}}
\end{center}
  \caption{An illustration of the classification and LDA projection. The projected data by LDA have greater distance between class centers than the ones by classification. Obviously, the nearest neighbors are not in the same category in (a).}
  \label{lda} 
\end{wrapfigure}

Unlike the unsupervised methods, supervised hashing directly resorts to the labels that indicate similar or dissimilar items.
Hence, it is natural to formulate the hashing problem as a classification task, where the projected codes of the same category are classified as the similar items.
CCA-ITQ~\cite{gong2013iterative} extends the ITQ into the supervised scenario by maximizing the correlation between features and corresponding labels, then the learned projection matrix is employed for generating codes.
\emph{Supervised Discrete Hashing} (SDH)~\cite{shen2015supervised} and \emph{Fast SDH} (FSDH)~\cite{koutaki2016fast} directly generate the hash codes within the linear multi-class classification framework, which shows considerable performance compared with before. However, these methods suffer from the weakness of shallow model in nonlinear modeling.
Recently, as the deep networks, especially ~\emph{Convolution Neural Network} (CNN), show effectiveness in feature learning and nonlinear modeling, more attention are paid to the deep supervised hashing.
To be specific, the early works usually employ the siamese network (i.e., two stream networks but share the same parameters)~\cite{chopra2005learning} to learn if two items are similar or not~\cite{li2015feature,zhu2016deep}.
Considering that the pair-loss is the indirect supervision over networks, why not directly utilizing the label of class information? Yang et al.\cite{yang2018supervised} formulate the deep hashing learning as a multi-class classification by minimizing the cross-entropy error. Li et al.~\cite{li2017deephashing} combine the pair-label and direct class label to supervise the hashing network, similar framework can also be found in~\cite{yao2016deep}.

Actually, if we rethink the meaning of the original hashing objective for similarity retrieval, we can find that the supervised classification fashion does not entirely meet the hashing requirements.
Hashing hopes to retrieve similar points in the Hamming space, which means the similar items should be close to each other but away from the dissimilar ones.
Although the nearby points of different categories are successfully classified by the learned hyperplane, they are still considered as the similar items according to the nearest neighbor searching.
Fortunately, LDA exactly fits the hashing objective, which aims to minimize intra-class covariance and maximize inter-class covariance simultaneously\cite{mika1999fisher}.
A simple illustration of such projection can be found in Fig.~\ref{lda}. It is easy to find that LDA is more suitable for retrieving similar points in the hashing task compared with the classical classification objective.
However, the existing \emph{LDA Hashing} (LDAH)~\cite{strecha2012ldahash} relies on the linear projection over hand-craft features, which limits its capacity in learning effective hashing function.

In this paper, to overcome the weaknesses of LDAH, we propose to learn hashing function under a kind of revised LDA objective but based on the effective deep feature learning.
That is, CNNs show noticeable effectiveness in learning semantic features, especially for the image messages~\cite{lecun2015deep}.
Meanwhile, the efficient nonlinear projection of deep models also provide possibilities to encode the original features into proper distribution in the low dimensional space~\cite{salakhutdinov2009semantic}.
These two properties exactly remedy the defects of the original LDA hashing.
However, simply performing the LDA objective over the networks is difficult to solve, as it has to be faced with the complicated eigenvalue decomposition for each mini-batch data in every epoch.
In this work, the revised LDA objective is transformed into a simple least square problem, which can be easily solved by the off-the-shelf optimizer, such as stochastic gradient descent.
To directly generate the hashing codes without additional binarization, a truly end-to-end hashing network is proposed by utilizing an adaptive binary activation function.
The proposed method is evaluated on three large-scale benchmark datasets, and beats several state-of-the-art methods on various metrics.



\section{LDA Hashing Revisited}
LDA hashing targets to project the high dimensional data ${x_i} \in {R^d}$ into the $r$-bits ($d \gg r$) short binary codes, while jointly minimizing the intra-class covariance and maximizing the inter-class covariance~\cite{strecha2012ldahash}.
To be specific, given the training dataset $\xi = \left\{ {\left( {{x_i},{c_i}} \right)|{x_i} \in {R^d},i = 1,2,...,n} \right\}$, where each data point ${x_i}$ corresponds to one class label ${c_i} \in \left\{ {1,2,...,c} \right\}$, and let ${{\bar x}} = \frac{1}{n}\sum\nolimits_{i = 1}^n {{x_i}} $ and ${{\bar x}_i} = \frac{1}{{{n_i}}}\sum\nolimits_{{x_j} \in {\chi _i}} {{x_j}}$ denote the global mean and specific-class mean (e.g., the $i$-class), respectively. Then the intra-class scatter matrix $S_w$,  the inter-class scatter matrix $S_b$, and the total-class scatter matrix $S_t$ can be written as
\begin{equation}\label{lda1}
\left\{ {\begin{array}{*{20}{c}}
{{S_w}{\rm{ = }}\sum\limits_{i = 1}^c {\sum\limits_{x \in {\chi _i}} {\left( {x - {{\bar x}_i}} \right){{\left( {x - {{\bar x}_i}} \right)}^T}} } }\\
{{S_b} = \sum\limits_{i = 1}^c {{n_i}\left( {{{\bar x}_i} - \bar x} \right){{\left( {{{\bar x}_i} - \bar x} \right)}^T}}  {\rm{~~~}}}\\
{{S_t} = \sum\limits_{i = 1}^n {\left( {{x_i} - \bar x} \right){{\left( {{{\bar x}_i} - \bar x} \right)}^T}} {\rm{~~~~~~~}}}
\end{array},} \right.
\end{equation}
where ${S_t} = {S_w} + {S_b}$.

Up to this point, LDA hopes to learn an optimal $W$ that makes the projected data points satisfy the above objective, i.e., minimizing the intra-class and maximizing the inter-class covariance. Hence, the classical LDA objective can be formulated as~\cite{jia2009trace},
\begin{equation}\label{lda2}
\mathop {\max }\limits_W Tr\left( {{{\left( {{W^T}{S_w}W} \right)}^{ - 1}}{W^T}{S_b}W} \right).
\end{equation}
Generalized eigenvalue decomposition can be used to efficiently solve this problem, i.e., ${S_b}{w_k} = {{\lambda}_k} {S_w}{w_k}$, where ${{\lambda}_k}$ and $w_k$ are the $k$-th largest generalized eigenvalue and corresponding eigenvector. Then, the projection matrix $W$ can be constituted by the eigenvectors with maximal $r$ eigenvalues. Hence, the projected low dimensional representation for data ${x_i}$  can be obtained by
\begin{equation}\label{lda3}
{h_i} = W^T \cdot {x_i}.
\end{equation}
However, due to the required binary constraint, hashing is obviously different from traditional dimensionality reduction. Directly imposing the binary constraint into the LDA objective will make it become a NP-hard problem. A common relaxation strategy is performing the binarization based on threshold selection~\cite{strecha2012ldahash},
\begin{equation}\label{lda4}
{b_i} = {\rm{sign}}(W^T \cdot {x_i} + t),
\end{equation}
where $\rm{sign}$ is the sign function and $t$ is the threshold.

Similar with the classical LDA method, LDA hashing also suffers from the so-called ``small sample size '' (SSS) problem~\cite{lu2005regularization}. That is, the original limited data points usually lie in high-dimensional space (e.g., 128-D in SIFT space, 4096-D in deep space), which results in sparse space when the data points are not enough for exploring the data structure as expected. Hence, the intra or inter-class covariance are usually not invertible when solving Eq.~\ref{lda2}. The common strategy is adding a regularization term to the original LDA objective as follows~\cite{friedman1989regularized},
\begin{equation}\label{lda5}
\mathop {\max }\limits_W Tr\left( {{{\left( {{W^T}{(S_w + \mu I)}W} \right)}^{ - 1}}{W^T}{S_b}W} \right),
\end{equation}
where $\mu$ is the regularization parameter.
Although the LDA objective exactly meets the requirements of hashing technique, the complicated feature distribution makes it difficult to learn effective low-dimensional representation by employing simple linear projection. Hence, LDA hashing has not shown noticeable performance before.
Even so, how about performing nonlinear feature learning to make it more discriminative and adaptable to the LDA objective?


\section{Deep LDA Hashing}
To overcome the weakness of original linear LDAH, \emph{Deep LDA Hashing} (DLDAH) actually does not learn the projection matrix over features but aims to directly learn efficient binary representations from raw image via multiple layers of nonlinear projection (i.e., CNN), which should satisfy the LDA objective. Hence, the LDA objective is directly performed over the final layers of CNN networks.
Formally, let $X \in {R^{d \times n}}$ and $Y \in {\left\{ {0,1} \right\}^{n \times c}}$ denote the final feature representation and corresponding one-hot labels, respectively.
Then the output feature $X$ of network $f$ should be encouraged to shrink the intra-class covariance and enlarge the inter-class covariance.
For better expression, define ${A^t} = \frac{1}{n}{11^T}$ and
$    A_{ij}^w = \left\{ {\begin{array}{*{20}{c}}
{\frac{1}{{{n_{{c_i}}}}}{\rm{~~~~~~}}{c_i} = {c_j}{\rm{~~~~~~}}}\\
{0{\rm{~~~~~~~~otherwise~~}}}
\end{array}}, \right.$ then the intra-class scatter matrix $S_w$ and the total-class scatter matrix  $S_t$ can be written as
\begin{equation}\label{dlda2}
    {S_w}{\rm{ = }}X\left( {I - {A^w}} \right){\left( {I - {A^w}} \right)^T}{X^T},
\end{equation}
and
\begin{equation}\label{dlda3}
{S_t} = X\left( {I - {A^t}} \right){\left( {I - {A^t}} \right)^T}{X^T}.
\end{equation}

As ${S_t} = {S_b} + {S_w}$, then the inter-class scatter matrix $S_b$ can be derived as follows,
\begin{equation}\label{dlda4}
   \begin{array}{l}
{S_b} = {S_t} - {S_w}\\
{\rm{=  }}X\left( {I - {A^t}} \right)\left( {I - {A^t}} \right){X^T} - X\left( {I - {A^w}} \right)\left( {I - {A^w}} \right){X^T}\\
{\rm{     =  }}X\left( {I - {A^t}} \right){X^T} - X\left( {I - {A^w}} \right){X^T}\\
{\rm{     =  }}X\left( {{A^w} - {A^t}} \right){X^T}\\
{\rm{     =  }}XH\left( {{A^w} - {A^t}} \right)H{X^T}\\
{\rm{     =  }}XH{A^w}H{X^T}\\
{\rm{     =  }}XHY{\left( {{Y^T}Y} \right)^{ - 1}}{Y^T}H{X^T}
\end{array}
\end{equation}
where $H = \left( {I - {A^t}} \right)$ is the centering matrix. Hence, the objective of DLDAH can be written as
\begin{equation}\label{dlda6}
    \mathop {\max }\limits_f Tr\left( {S_t^{ - 1}{S_b}} \right),
\end{equation}
where $f$ stands for the deep networks with multiple layers of nonlinear projection.
Note that, as the total-class scatter matrix ${S_t} = {S_b} + {S_w}$, Eq.\ref{dlda6} is equivalent to maximizing ${S_b}$ and minimizing ${S_w}$~\cite{ye2007least}.
Compared with Eq.\ref{lda2},  the linear projection matrix $W$ is \textbf{replaced} with multiple layers of nonlinear projection $f$, which means the proposed model has stronger ability in regularizing the learned features to the hashing objective.
However, the matrix $S_t$ may be not invertible due to the SSS problem, an additional regularization term is considered and Eq.\ref{dlda6} becomes,
\begin{equation}\label{dlda7}
    \mathop {\max }\limits_f Tr\left( {{{\left( {{S_t} + \mu I} \right)}^{{\rm{ - }}1}}{S_b}} \right).
\end{equation}

Although the revised objective of deep network currently meets the requirements of hashing technique, we have to be faced with two intractable problems when training the deep networks with Eq.\ref{dlda7}.
On the one hand, due to the huge training data and amounts of trainable parameters, network optimization is iteratively performed with small batches of the database. However, such optimization strategy is not entirely suitable for the revised LDA objective, i.e., Eq.\ref{dlda7}. This is because all the scatter matrices are computed based on the whole training data, only a small batch of them can not exactly reflect the complex distribution of feature points.
On the other hand, as introduced in Sec.2, generalized eigenvalue decomposition is usually adopted to solve the LDA objective, whose time complexity is $O\left( {{d^3}} \right)$. When performing such optimization within the batch-strategy of deep networks, the complexity becomes $O\left( {{d^3} \cdot m \cdot iter} \right)$, where $d$ is the feature dimension, $m$ is the number of batches of each epoch and $iter$ is the number of epoches. It becomes impractical to perform eigenvalue decomposition over the deep features under such huge time complexity.
Hence, these two intractable problems make it difficult to directly optimize Eq.\ref{dlda7}.
Fortunately, according to Theorem 1, we can convert the original problem into a well-developed framework and solve the two problems simultaneously.

\noindent \textbf{Theorem 1.} \emph{Minimizing the linear regression of least square is equivalent to maximizing the LDA objective in Eq.\ref{dlda7}\footnote{The proof is available in the supplementary material.}.}

Hence, Eq.\ref{dlda7} can be transformed in,
\begin{equation}\label{dlda8}
    \mathop {\min }\limits_{f,W,b} \left\| {{X^T}W + 1{b^T} - \widetilde Y} \right\|_F^2 + \mu \left\| W \right\|_F^2,
\end{equation}
where $\widetilde Y = Y{\left( {{Y^T}Y} \right)^{ - {1 \mathord{\left/ {\vphantom {1 2}} \right. \kern-\nulldelimiterspace} 2}}}$, $W$ and $b$ are the newly introduced regression matrix and bias term, respectively.
It is easy to find that the aforementioned problems are solved automatically. The least square error does not extremely suffer from the problem of optimization based on mini-batch, and the time complexity for the regression layer is the same as standard fully connected layers. Hence, it becomes feasible to optimize the deep network with the proposed LDA objective.

Although the previous works~\cite{ye2007least, sun2010scalable, lee2015equivalence} have shown the equivalence between multivariate linear regression and classical LDA objective, i.e., Eq.\ref{lda2}, such equivalence is obviously different from the conditions in Theorem 1 and cannot be applied here.
On the one hand, the classical LDA objective has been revised for supervising deep models, where the linear projection matrix is replaced with multi-layers of nonlinear transformation. Such modification leads to the disparate least square objective, i.e., the newly introduced variable and different indicator matrix $\widetilde Y$.
On the other hand, the previous equivalence~\cite{ye2007least} requires rank($S_b$) + rank($S_w$) = rank($S_t$), which does not always hold for the high-dimensional and under-sampled data~\cite{ye2006computational}, i.e., the SSS condition. In contrast, our proof takes no assumption about the scatter matrix, therefore it is comfortable for deep supervision.

\subsection{Controllable Projection}
By comparing Eq.\ref{dlda7} and Eq.\ref{dlda8}, we can find that the original one-hot label $Y$ is changed into $\widetilde Y$ that considers the imbalance between different categories via ${\left( {{Y^T}Y} \right)^{ - {1 \mathord{\left/ {\vphantom {1 2}} \right. \kern-\nulldelimiterspace} 2}}}$. But how such variation affects the proposed LDA objective, and how about directly employing the original label $Y$ without the modification?
Hence, Eq.\ref{dlda8} is further explored to answer these questions.

To simplify the analysis, the original label $Y$ is adopted instead of $\widetilde Y$. Then, Eq.\ref{dlda4} becomes
\begin{equation}\label{cp1}
    {S_b}{\rm{ =  }}XHY{Y^T}H{X^T}.
\end{equation}
Substituting Eq.\ref{cp1} into Eq.\ref{dlda7},
\begin{equation}\label{cp2}
   \begin{array}{l}
Tr\left( {{{\left( {{S_t} + \mu I} \right)}^{ - 1}}{S_b}} \right)\\
 = Tr\left[ {{{\left( {{S_t} + \mu I} \right)}^{ - {1 \mathord{\left/
 {\vphantom {1 2}} \right.
 \kern-\nulldelimiterspace} 2}}}{{\left( {{S_t} + \mu I} \right)}^{ - {1 \mathord{\left/
 {\vphantom {1 2}} \right.
 \kern-\nulldelimiterspace} 2}}}XHY{Y^T}H{X^T}} \right]\\
 = Tr\left[ {{{\left( {{S_t} + \mu I} \right)}^{ - {1 \mathord{\left/
 {\vphantom {1 2}} \right.
 \kern-\nulldelimiterspace} 2}}}XHY \cdot {Y^T}H{X^T}{{\left( {{S_t} + \mu I} \right)}^{ - {1 \mathord{\left/
 {\vphantom {1 2}} \right.
 \kern-\nulldelimiterspace} 2}}}} \right]\\
{\rm{ = }}\left\| {{{\left( {{S_t} + \mu I} \right)}^{ - {1 \mathord{\left/
 {\vphantom {1 2}} \right.
 \kern-\nulldelimiterspace} 2}}}XHY} \right\|_F^2.
\end{array}
\end{equation}
The original ratio trace becomes the Frobenius norm of the projected centering feature matrix. It can be further written w.r.t. specific-classes as follows,
\begin{equation}\label{cp3}
\begin{array}{l}
\left\| {{{\left( {{S_t} + \mu I} \right)}^{ - {1 \mathord{\left/
 {\vphantom {1 2}} \right.
 \kern-\nulldelimiterspace} 2}}}XHY} \right\|_F^2\\
{\rm{ = }}\left\| {{{\left( {{S_t} + \mu I} \right)}^{ - {1 \mathord{\left/
 {\vphantom {1 2}} \right.
 \kern-\nulldelimiterspace} 2}}}\left[ {{n_1}\left( {{{\bar x}_1} - \bar x} \right),...,{n_c}\left( {{{\bar x}_c} - \bar x} \right)} \right]} \right\|_F^2\\
 = \sum\limits_{i = 1}^c {n_i^2\left\| {{{\left( {{S_t} + \mu I} \right)}^{ - {1 \mathord{\left/
 {\vphantom {1 2}} \right.
 \kern-\nulldelimiterspace} 2}}}{{\bar x}_i} - {{\left( {{S_t} + \mu I} \right)}^{ - {1 \mathord{\left/
 {\vphantom {1 2}} \right.
 \kern-\nulldelimiterspace} 2}}}\bar x} \right\|_2^2}
\end{array}.
\end{equation}
Eq.\ref{cp3} means that the class mean and the global mean are normalized by the total-class scatter matrix, and the inter-class covariance is maximized by enlarging their Euclidean distance. Note that the label $Y$ performs as a parameter to amplify the inter-class distance according to the number of items of each class, which prompts these class-centers to be as far as possible.
When $Y$ is replaced with $\widetilde Y$,  Eq.~\ref{dlda7} can be re-written as
\begin{equation}\label{cp4}
\sum\limits_{i = 1}^c {n_i \left\| {{{\left( {{S_t} + \mu I} \right)}^{ - {1 \mathord{\left/
 {\vphantom {1 2}} \right.
 \kern-\nulldelimiterspace} 2}}}{{\bar x}_i} - {{\left( {{S_t} + \mu I} \right)}^{ - {1 \mathord{\left/
 {\vphantom {1 2}} \right.
 \kern-\nulldelimiterspace} 2}}}\bar x} \right\|_2^2}.
\end{equation}
Compared with Eq.\ref{cp3}, the parameter of $n_i^2$ becomes $n_i$ in Eq.\ref{cp4}. Then, the inter-class covariance becomes smaller than the original one.
Hence, for simplicity and enlarging the inter-class distance, the label matrix $Y$ is adopted instead of  $\widetilde Y$ in Eq.\ref{dlda8} in this paper.



\subsection{End-to-end hashing network}
Hashing technique requires the low-dimensional representation to be binary. As discussed in Sec.2, it becomes a NP-hard problem when solving the LDA objective with the binary constraint. The common strategy is to perform the binarization after projecting the original data into the low-dimensional space~\cite{strecha2012ldahash}. However, such separated two-stage methods may destroy the learned space and result in sub-optimal hashing codes~\cite{li2017deep}.
Although the recent deep hashing methods propose to reduce the quantization loss when optimizing the networks, the activations of the hashing layer are still not binary. This is because the sign function have no gradient, and most works have to use the tanh function as the approximated binary activation within the network~\cite{zhu2016deep,li2017deephashing}.

Inspired by \cite{li2017deep}, a learnable parameter is introduced into the original tanh function, which can avoid the tiny gradient when faced with large input to some extent and project the input into the discrete domain
of $\left\{1, -1 \right\}$. Such activation function is named as \emph{Adaptive Tanh} (ATanh) and written as
\begin{equation}\label{net1}
    {b_i} = \tanh \left( {\alpha {x_i}} \right) + \nu \left\| {{\alpha ^{ - 1}}} \right\|_2^2,
\end{equation}
where $\nu$ is the regularization constant. By minimizing the regularization term of $\left\| {{\alpha ^{ - 1}}} \right\|_2^2$, $\alpha$ can gradually increase so that the final activations approach the sign function and have the ability to generate binary codes. To directly perform the LDA objective over the binary representation and generate efficient hashing codes within DLDAH, the ATanh function is performed over the last layer but before the regression layer of LDA, whose activations constitute the hashing layer.
Different from the previous unimodal hashing network, there is no extra quantization loss within such end-to-end hashing net, hence it shows stronger capacity in learning efficient codes.
Meanwhile, as the LDA regression, i.e. Eq.~\ref{dlda7}, is directly performed over the binary values, the generated hashing codes within the same class enjoy smaller hamming distance while different classes take larger distance.


\section{Experiments}
\subsection{Dataset}

\noindent{\textbf{MNIST}~\cite{lecun1998gradient}} consists of 70,000 $28\times28$ grey-scale images associated with digits from 0  to  9. For the unsupervised methods, we randomly select 100 samples for each digit as the query images and all the rest ones constitute the training set. For the supervised methods, 500 images per class are selected as the training set. For all the methods, all the samples except the testing ones construct the retrieval gallery set.

\noindent{\textbf{CIFAR-10}~\cite{krizhevsky2009learning}} consists of 60,000 32$\times$32 RGB images that are divided into 10 object categories and 6,000 images per category. Following the settings in \cite{wang2016deep}, 100 image samples per class are randomly selected as the testing items and the training set consists of the remaining samples for the unsupervised methods. As for the supervised ones, 500 samples are uniformly selected from each category.

\noindent{\textbf{ImageNet}~\cite{deng2009imagenet}} is a benchmark dataset built for \emph{Large Scale Visual Recognition Challenge} (ILSVRC). In this paper, the ILSVRC2012 version is chosen for evaluation. It consists of about 1.3 million images that are labeled into 1,000 categories, where one image only corresponds to one label. For efficiency, we select the most frequent 100 categories. Following \cite{cao2017hashnet}, 100 images per category form the training set, and the 50 validation images of each class constitute the testing set. While for the unsupervised methods, we employ all the images of about 130K for training. And all of the images excluding query sets are also treated as the retrieval gallery.

\subsection{Evaluation}
In order to effectively evaluate the proposed model, the representative data-independent and data-dependent hashing methods are considered. They are the unsupervised methods LSH, SH, and ITQ, the conventional supervised methods LDAH, SDH, and FSDH, and the state-of-the-art deep hashing methods DHN~\cite{zhu2016deep}, DTSH~\cite{wang2016deep}, HashNet~\cite{cao2017hashnet}, and DSDH~\cite{li2017deephashing}.
To comprehensively compare all these methods, both the global and local evaluation are considered, i.e., Hamming ranking and hash lookup.
Specifically, ~\emph{Mean Average Precision} (MAP) is employed for computing the quality of the whole retrieved sequence according to the Hamming distance to a query. While for the hash lookup, only the top retrieved samples are considered, i.e., the $\left\{Recall, F-measure \right\}$ score is calculated based on a Hamming ball of radius 2 to a query~\cite{liu2014discrete, li2017large}.

\subsection{Set up}
As the proposed model is an extension of the classical LDA on deep models, it is necessary to compare with other deep hashing methods with the same basic network for fairness.
Following \cite{wang2016deep,li2017deephashing}, we adopt the VGG-F model\footnote{http://www.vlfeat.org/matconvnet/pretrained/} pretrained on the ImageNet as the deep architecture, so are the other deep models. The provided results are obtained by running the source code provided by the authors. As the high-level CNN representations better reflect the semantic of images, they are also employed as the features for all the conventional shallow models. And the hype-parameter $\nu$ of ATanh follows the empirical value of 0.001 in \cite{li2017deep}, and $\mu$ is set to $0.0005$.


\begin{table*}[t]
\centering
\small
\newcolumntype{C}[1]{>{\centering}p{#1}}
\caption{\label{Table2}The comparison results of different hashing methods in MAP on CIFAR-10 and ImageNet.}
\vskip 0.15in
\begin{tabular}{C{2cm}|C{0.8cm}C{0.8cm}C{0.8cm}C{0.8cm}C{0.8cm}|C{0.8cm}C{0.8cm}C{0.8cm}C{0.8cm}}
\hline
  Dataset &  \multicolumn{5}{c|}{CIFAR-10} &  \multicolumn{4}{c}{ImageNet}
\tabularnewline\cline{1-10}

 Code $\#$bits                        & $8$ & $16$  & $32$ & $64$ &  $128$ &           $8$ & $16$ & $32$ & $48$
\tabularnewline
\hline
  LSH     & 0.1374&	0.1403	&	0.1693&0.1643&	0.2062	  & 0.0211	&0.0260	&0.0475	&0.0815	           \tabularnewline

  SH     &0.1920&	0.1810	&0.1700	&0.1690&	0.1696	      & 0.0815&	0.1118	&0.1578&	0.1869  \tabularnewline

   ITQ    &0.2308	&0.2480	&0.2540&	0.2750	&0.2892      &  0.1163	&0.1883	&0.2723	&0.3152\tabularnewline

\hline
   LDAH     & 0.1677&0.1408	&0.1243	&0.1162	&0.1120        &   0.0659	&0.1122&	0.1946&	0.2540 \tabularnewline

   SDH     &0.4881&	0.5516	&0.5807	&0.5925	&0.6067	        & 0.1840	&0.3172&	0.4088	&0.4514	\tabularnewline

   FSDH  &N/A&	0.5616	&0.5616&	0.5616	&0.5616      &N/A &N/A &N/A &N/A  \tabularnewline

 \hline

   DHN       &0.5750	&0.6688	&0.7004	&0.7027	&0.7078  &	0.2388&	0.3650	&0.4475&	0.4884 \tabularnewline

   HashNet  &0.5813&	0.6733	&0.7128	&0.6809	&0.7074  &\underline{0.2439}&	\underline{0.3819}&	0.4756	&0.5262		\tabularnewline

   DTSH      &0.6214	&0.7094	&0.7613&	0.7743&	0.7213	&  0.2118	&0.3497	&0.4433&	0.5035  \tabularnewline

   DSDH     & \underline{0.6774}&	\underline{0.7261}&\underline{0.7623}&	\underline{0.7905}&	\underline{0.7502}&   0.2313&	0.3720&\underline{0.4810}	&\underline{0.5326}\tabularnewline

 \hline
  DLDAH    &\textbf{0.7006}	&\textbf{0.7500}	&\textbf{0.7724}&\textbf{0.7964}&\textbf{0.8002}   & \textbf{0.2930}&\textbf{0.3953}&	\textbf{0.5208}	&\textbf{0.5624}		   \tabularnewline

  \hline

\end{tabular}
\vskip -0.1in
\end{table*}

\begin{figure*}[t]
\centering
\includegraphics[width=14cm]{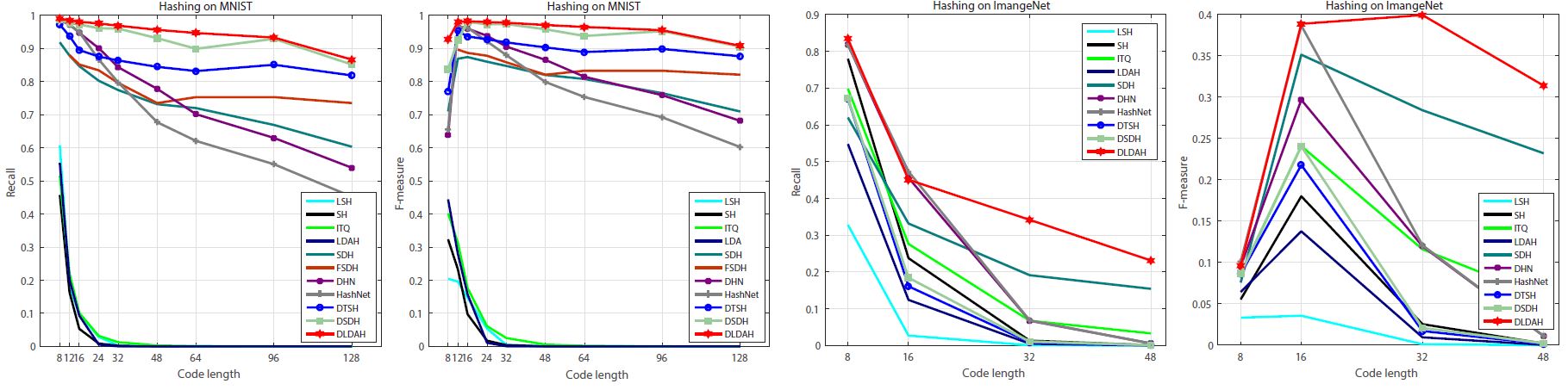}\\
\caption{The hash lookup performance of different hashing methods on MNIST and ImageNet dataset with varying code lengths.}\label{lookup1}
\end{figure*}

\begin{figure*}[t]
\centering
\includegraphics[width=14cm]{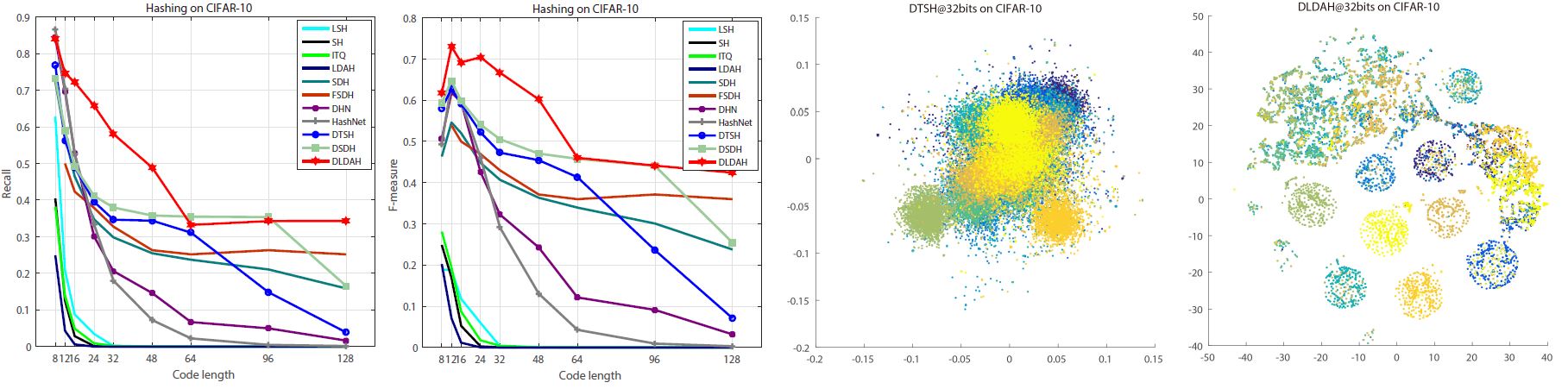}\\
\caption{The hash lookup performance and t-SNE visualization on CIFAR-10 dataset with varying code lengths.}\label{lookup2}
\end{figure*}

\subsection{Results and analysis}
\noindent \textbf{Results on MNIST.}
The hash lookup results in F-measure and Recall are shown in Fig.~\ref{lookup1} and the proposed DLDAH outperforms all the other methods on all the code lengths.
In both Recall and F-measure, the performance of all the methods decrease with the increasing code length, especially the unsupervised ones. Such phenomenon results from the more sparse hamming space~\cite{li2017large}. However, although DLDAH also suffers from the same problem, it still remains stable and substantial superiority over the other ones.

\noindent \textbf{Results on CIFAR-10.}
The Hamming ranking performance is shown in Table \ref{Table2}, and the proposed DLDAH shows the best results, especially better than the classical LDAH by 70 points.
Besides, HashNet is an extension of the deep model of DHN, which employs a fixed sequence of parameterized tanh as the activation function for training the network step by step. As it can directly generate the binary codes after training the network, the codes become more efficient. By contrast, DLDAH adopts the learnable parameterized tanh instead of the fixed tanh in HashNet, which could adaptively learn more effective hashing function within the end-to-end  network, hence, it shows much improvement than the other deep models.

Fig.~\ref{lookup2} shows the hash lookup performance in terms of F-measure and recall.
In Fig.~\ref{lookup2}, the conventional shallow models of SDH and FSDH outperform some deep hashing methods on both metrics, especially when the code is longer than 32 bits. The generated hashing codes by deep models are not as efficient as expected. This is because it is difficult to optimize deep hashing networks under complex code constraints, while it is easy for the shallow linear models. In contrast, due to the efficient LDA supervision, the proposed DLDAH still enjoys reliable codes when the codes become longer.
On the other hand, all of these methods suffer from the similar conditions on MNIST but even worse, however, DLDAH remains stable when the code is longer than 64 bits.
Besides, although DLDAH performs a litter worse than DSDH at 64 bits in recall, it shows much better F-measure and outperforms DSDH on all the lengths.

Different from the quantitative evaluation above, we also visualize the learned hashing codes by embedding them into 2D space via t-SNE~\cite{maaten2008visualizing}. The state-of-the-art method of deep hashing based on classification, i.e., DTSH, is chosen for comparison. DTSH aims to make the query close to the positive items but away from the dissimilar ones. As shown in Fig.~\ref{lookup2}, most of the hashing codes learned by  DTSH are actually mixed together, although some categories are successfully classified.
Yet, the codes of DLDAH show distinctly discriminative structure for efficient similarity retrieval, which benefit from the efficient supervision of LDA.
Such structure provides possibilities to achieve both high Precision and Recall, as shown above.
Even so, some samples are still not successfully separated from the other ones by DLDAH in Fig.~\ref{lookup2}.
This could be because the deep features of different categories have similar semantic, which will be considered for further study.

\noindent \textbf{Results on ImageNet.}
As FSDH requires that the number of bits should be greater than the number of classes but there is 100 categories in the ImageNet dataset, its results are not shown in this part.
Meanwhile, due to the time constraint, we only show the results with limited code lengths.
As shown in Table~\ref{Table2}, DLDAH outperforms all the other methods by 1-5 points in MAP.
Further, we also show the hash lookup performance in Fig.~\ref{lookup1}.
To maintain the performance when faced with much more categories in the ImageNet dataset, the hashing codes should be more efficient within specific Hamming ball.
However, the hashing function are not effectively learned by most deep methods as expected, whose performances decrease rapidly with the increasing code length.
Such phenomenon also comes from the inefficient supervision of classification, as shown in Fig.~\ref{lookup2}.
However, DLDAH can separate these different categories in the Hamming space to some extent and learn discriminative hashing codes for efficient retrieval.

\section{Conclusion}
In this paper, we show that the conventional supervised hashing methods based on classification are not entirely suitable for hashing learning.
To perform efficient similarity retrieval, a revised LDA objective is proposed to perform over the learned features by deep network, which encourages the similar samples to have small covariance while the dissimilar ones to have large covariance. And the deep hashing network with adaptive binary activation is optimized w.r.t the simple least square regression, which is proved to be equivalent to the original LDA objective.
Such model provides a brand new viewpoint in analyzing, learning, and transforming the data structure, and the projected binary codes show better discriminative structure for hashing retrieval.
Several deep hashing methods based on classification are defeated and considerable improvements are confirmed on three benchmark datasets.

\bibliography{dldah}
\bibliographystyle{ieee}

\end{document}